\documentclass{article}

\PassOptionsToPackage{numbers, compress}{natbib}
\usepackage[final]{neurips_2021}

\usepackage[utf8]{inputenc} % allow utf-8 input
\usepackage[T1]{fontenc}    % use 8-bit T1 fonts
\usepackage{hyperref}       % hyperlinks
\usepackage{url}            % simple URL typesetting
\usepackage{booktabs}       % professional-quality tables
\usepackage{amsfonts}       % blackboard math symbols
\usepackage{nicefrac}       % compact symbols for 1/2, etc.
\usepackage{microtype}      % microtypography
\usepackage{xcolor}         % colors

\usepackage{color}
\usepackage{multirow}
\usepackage{array}
\usepackage{graphicx}
\usepackage{amsmath}
\usepackage{amssymb}
\usepackage{lipsum}
\usepackage{bbm}
\usepackage{bm}
\usepackage{wrapfig}
\usepackage{epstopdf}
\usepackage{subcaption}
\usepackage{comment}

\newcommand\independent{\protect\mathpalette{\protect\independenT}{\perp}}
\def\independenT#1#2{\mathrel{\rlap{$#1#2$}\mkern2mu{#1#2}}}
\newcommand{\tabincell}[2]{\begin{tabular}{@{}#1@{}}#2\end{tabular}}

\title{DP-SSL: Towards Robust Semi-supervised Learning with A Few Labeled Samples}

\author{%
Yi Xu\textsuperscript{\rm 1}\quad
Jiandong Ding\textsuperscript{\rm 2}\quad
Lu Zhang\textsuperscript{\rm 1}\quad
Shuigeng Zhou \textsuperscript{\rm 1}\thanks{Corresponding author.}\\
\textsuperscript{\rm 1}Shanghai Key Lab of Intelligent Information Processing, \\
and School of Computer Science, Fudan University, China\\
{\textsuperscript{\rm 2}Alibaba Group}\\
{\tt \{yxu17, jdding, l\_zhang19, sgzhou\}@fudan.edu.cn} \\
}

\begin{document}
	
	\maketitle
	\begin{abstract}
		The scarcity of labeled data is a critical obstacle to deep learning. Semi-supervised learning (SSL) provides a promising way to leverage unlabeled data by pseudo labels. However, when the size of labeled data is very small (say a few labeled samples per class), SSL performs poorly and unstably, possibly due to the low quality of learned pseudo labels.
		In this paper, we propose a new SSL method called DP-SSL that adopts an innovative data programming (DP) scheme to generate probabilistic labels for unlabeled data. Different from existing DP methods that rely on human experts to provide initial labeling functions (LFs), we develop a multiple-choice learning~(MCL) based approach to automatically generate LFs from scratch in SSL style.
		With the noisy labels produced by the LFs, we design a label model to resolve the conflict and overlap among the noisy labels, and finally infer probabilistic labels for unlabeled samples. 
		Extensive experiments on four standard SSL benchmarks show that DP-SSL can provide reliable labels for unlabeled data and achieve better classification performance on test sets than existing SSL methods, especially when only a small number of labeled samples are available. Concretely, for CIFAR-10 with only 40 labeled samples, DP-SSL achieves 93.82\% annotation accuracy on unlabeled data and 93.46\% classification accuracy on test data, which are higher than the SOTA results.
	\end{abstract}
	
	\begin{comment}
	\begin{figure}[t]
	\begin{center}
	\fbox{\rule{0pt}{2in} \rule{0.9\linewidth}{0pt}}
	%\includegraphics[width=0.8\linewidth]{egfigure.eps}
	\end{center}
	\caption{Motivation}
	\label{fig:itro}
	\end{figure}
	\end{comment}
	
	\section{Introduction}
	\label{sec:introduction}
	\begin{wrapfigure}{R}{0.4\textwidth}
		\includegraphics[width=0.4\textwidth]{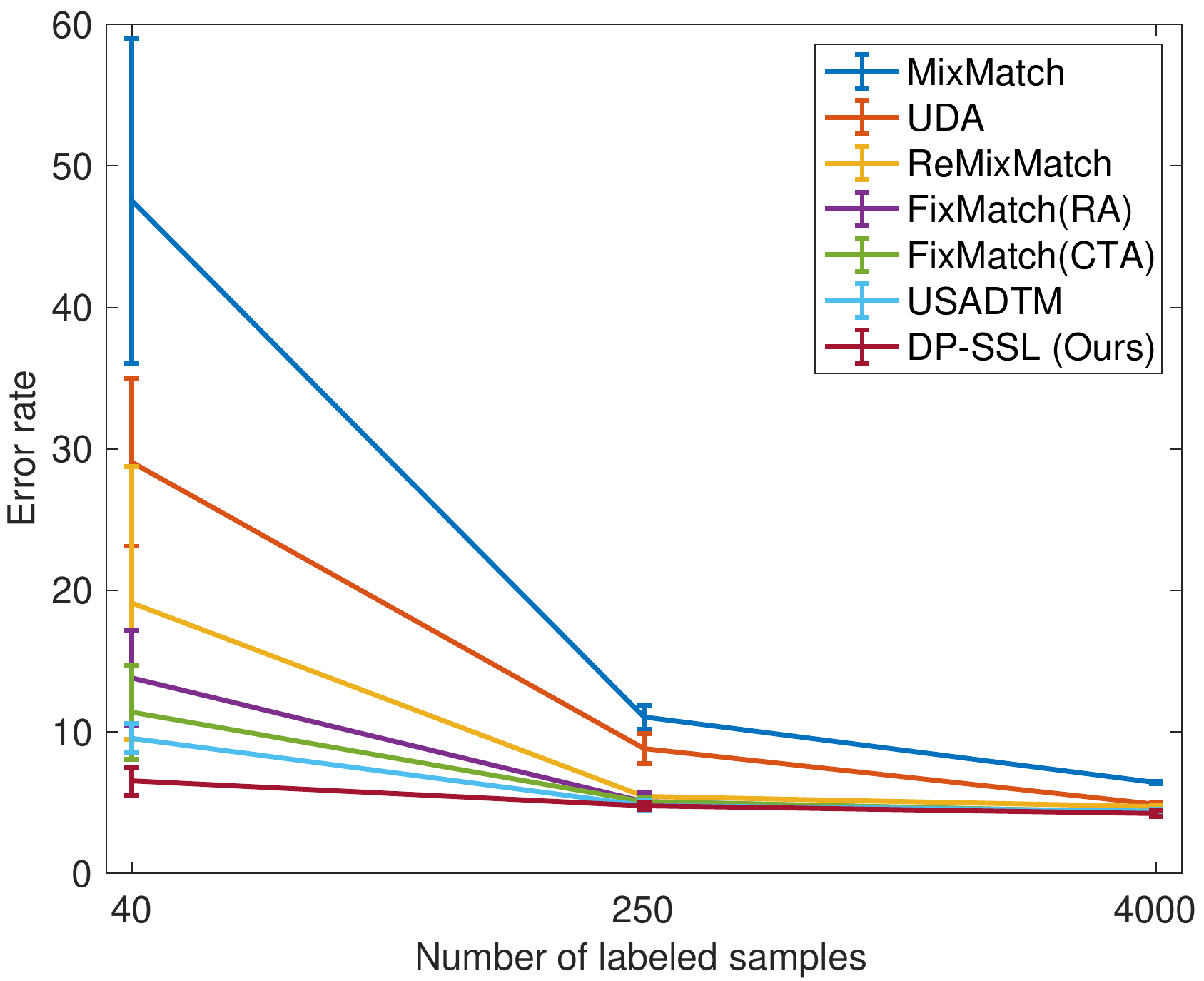}
		\caption{Error rate \emph{vs}. \#labeled samples (CIFAR-10). Results of existing methods are from the original papers. When only 40 labeled samples are given, all existing SSL methods are substantially degraded and more unstably, while our method is still effective and robust.}
		\label{fig:intro}
	\end{wrapfigure}
	The de-facto approaches to deep learning achieve phenomenal success with the release of huge labeled datasets. However, large manually-labeled datasets are time-consuming and expensive to acquire, especially when expert labelers are required. Nowadays, many techniques are proposed to alleviate the burden of manual labeling and help to train models from scratch, such as active learning~\cite{gao2020consistency}, crowd-labeling~\cite{vondrick2013efficiently}, distant supervision~\cite{yao2021visual},
	semi~\cite{sohn2020fixmatch}/weak~\cite{oquab2015object}/self-supervision~\cite{chen2020simple}.
	Among them, semi-supervised learning (SSL) is one of the most popular techniques to cope with the scarcity of labeled data. Two major strategies of SSL are pseudo labels~\cite{lee2013pseudo} and consistency regularization~\cite{sajjadi2016regularization}. Pseudo labels~(also called self-training~\cite{zoph2020rethinking}) utilize a model's predictions as the labels to train the model again, while consistency of regularization forces a model to make the same prediction under different transformations. However, when the size of labeled data is small, SSL performance degrades drastically in both accuracy and robustness.  Fig.~\ref{fig:intro} shows the change of prediction error rate with the number of labeled samples of CIFAR-10. When the number of labeled samples reduces from 250 to 40, error rates of major existing SSL methods increase from {4.74}\% (USADTM) to {36.49}\% (MixMatch). One possible reason of performance deterioration is the quality degradation of learnt pseudo labels when labeled data size is small. Therefore, in this paper we address this problem by developing sophisticated labeling techniques for unlabeled data to boost SSL even when the number of labeled samples is very small (e.g. a few labeled samples per class).

Recently, \emph{data programming}~(DP) was proposed as a new paradigm of weak supervision~\cite{ratner2016data}. In DP, human experts are required to transform the decision-making process into a series of small functions~(called \emph{labeling functions}, abbreviated as LFs), thus data can be labeled programmatically. 
	Besides, a label model is applied to determining the correct labels based on consensus from the noisy and conflicting labels assigned by the LFs. Such a paradigm achieves considerable success in NLP tasks~\cite{awasthi2020learning,ratner2017snorkel, ratner2019training,chatterjee2020data}. In addition, DP has also been applied to computer vision tasks~\cite{chen2019scene,Hooper2021cutout}. 
	However, current DP methods require human experts to provide initial LFs, which is time-consuming and expensive, and it is not easy to guarantee the quality of LFs. Furthermore, LFs specifically defined for one task usually cannot be re-used for other tasks.
	
In this paper, we propose a new SSL method called DP-SSL that is effective and robust even when the number of labeled samples is very small. In DP-SSL,
an innovative data programming (DP) scheme is developed to generate probabilistic labels for unlabeled data. Different from existing DP methods, we develop a \emph{multiple-choice learning}~(MCL) based approach to automatically generate LFs from scratch in SSL style.
To remedy the over-confidence problem with existing MCL methods, we assign an additional option as \emph{abstention} for each LF. 
After that, we design a label model to resolve the conflict and overlap among the noisy labels generated by LFs, and infer a probabilistic label for each unlabeled sample. Finally, the probabilistic labels are used to train the end model for classifying unlabeled data.
Our experiments validate the effectiveness and advantage of DP-SSL. As shown in Fig.~\ref{fig:intro}, DP-SSL performs best, and only {1.76}\% increase of error rate when the number of labeled samples decreases from 250 to 40 in CIFAR-10. 

Note that the pseudo labels used in existing SSL methods are quite different from the probabilistic labels in DP-SSL, which may explain the advantage of DP-SSL over existing SSL methods. On the one hand, pseudo labels are ``hard'' labels that indicate an unlabeled sample belonging to a certain class or not, while probabilistic labels are ``soft'' labels that indicate the class distributions of unlabeled samples. Obviously, the latter should be more flexible and robust. On the other hand, pseudo labels are actually generated by a single model for all unlabeled samples, while probabilistic labels are generated from a number of diverse and specialized LFs (due to the MCL mechanism), which makes the latter more powerful in generalization as a whole.

In summary, the contributions of this paper are as follows:
%\begin{itemize}
1) We propose a new SSL method DP-SSL that employs an innovative data programming method to generate probabilistic labels for unlabeled data, which makes DP-SSL effective and robust even when there are only a few labeled samples per class.
2) We develop a multiple choice learning based approach to automatically generate diverse and specialized LFs from scratch for unlabeled data in SSL manner.
3) We design a label model with a novel potential and an unsupervised quality guidance regularizer to infer probabilistic labels from the noisy labels generated by LFs.
4) We conduct extensive experiments on four standard benchmarks, which show that DP-SSL outperforms the state-of-the-art methods, especially when only a small number of labeled samples are available, DP-SSL is still effective and robust.
%\end{itemize}

	\section{Related Work}
Here we briefly review the latest advances in multiple choice learning, semi-supervised learning, and data programming, which are related to our work. Detailed information is available in \cite{garcia2021distillation,van2020survey,yang2021survey,chen2020train,fu2020fast}.

	\subsection{Multiple Choice Learning}
	Multiple choice learning (MCL)~\cite{guzman2014efficiently} was proposed to overcome the low diversity problem of models trained independently in ensemble learning. For example, stochastic multiple choice learning~\cite{lee2016stochastic} is for training diverse deep ensemble models.
	However, a crucial problem with MCL is that each model tends to be overconfident, which results in poor final prediction.
	To solve this problem, \cite{lee2017confident} forces the predictions of non-specialized models to meet a uniform distribution, so that the final decision is summed over diverse outputs. \cite{tian2019versatile} proposes an additional network to estimate the weight of each specialist's output.
	In this paper, we develop an improved MCL based scheme to automatically generate diverse and specialized labeling functions (LFs) from scratch in an SSL manner. These LFs are used to generate preliminary (usually noisy) labels for unlabeled data.

	\subsection{Semi-supervised Learning}
	Semi-supervised learning (SSL) has been extensively studied in image classification~\cite{deng2009imagenet}, object detection~\cite{lin2014microsoft}, and semantic segmentation~\cite{cordts2016cityscapes}. Two popular SSL strategies for image classification are pseudo labels~\cite{lee2013pseudo} and consistency regularization~\cite{sajjadi2016regularization}. Pseudo-label methods generate artificial labels for some unlabeled images and then train the model with these artificial labels, while consistency regularization tries to obtain an artificial distribution/label and applied it as a supervision signal with other augmentations/views. These two strategies have been adopted by a number of recent SSL works~\cite{sohn2020fixmatch, sajjadi2016regularization, laine2016temporal, rasmus2015semi, berthelot2019mixmatch, tarvainen2017mean, miyato2018virtual, xie2020unsupervised, berthelot2019remixmatch,han2020unsupervised, li2020comatch, hu2021simple, zhang2021flexmatch}. For example, FixMatch~\cite{sohn2020fixmatch} proposes a simple combination of pseudo labels and consistency regularization. 
\cite{han2020unsupervised} employs unsupervised learning and clustering to determine the pseudo labels. In this paper, we propose a new SSL method that is effective and robust even when the size of labeled data is very small. Our method employs an innovative data programming alike method to automatically generate probabilistic labels for unlabeled data.
	
	\subsection{Data Programming} Data programming~\cite{ratner2016data} is a weak supervision paradigm proposed to infer correct labels based on the consensus among noisy labels from labeling functions (LFs), which are modules embedded with decision-making processes for generating labels programmatically.
Following the DP paradigm, Snorkel~\cite{ratner2017snorkel} and Snuba~\cite{varma2018snuba} were proposed as rapid training data creation systems. Their LFs are built with various weak supervision sources, like pattern regexes, heuristics, and external knowledge base etc.
	Recently, more works are reported in the literature~ ~\cite{awasthi2020learning,ratner2019training, chatterjee2020data, chen2019scene, Hooper2021cutout,fu2020fast,  heo2020inspector, pal2018adversarial, das2020goggles, boecking2021interactive, varma2017inferring, bach2017learning, varma2019learning, zhang2021weakly}. Among them, \cite{awasthi2020learning, ratner2019training,chatterjee2020data, varma2017inferring, bach2017learning, varma2019learning} focus on the adaption of label model in DP. For example, \cite{fu2020fast} aims to reduce the computational cost and proposes a closed-formed solution for training the label model. \cite{chen2019scene, Hooper2021cutout, heo2020inspector, pal2018adversarial, das2020goggles} apply DP to computer vision. Concretely, \cite{Hooper2021cutout, pal2018adversarial, das2020goggles} heavily rely on the pretrained models. \cite{heo2020inspector} combines crowdsourcing, data augmentation, and DP to create weak labels for image classification. \cite{chen2019scene} presents a novel view for resolving infrequent data in scene graph prediction training datasets via image-agnostic features in LFs.
	However, all these methods cannot be directly applied to training models from scratch with a small number of labeled samples. Thus, in this paper we extend DP by exploring both MCL and SSL to generate arbitrary labeling functions.

	\section{Method}
For a $C$-class SSL classification problem, assume that all training data $X$ are divided into labeled data $X_l$ and unlabeled $X_u$, and test data are denoted as $X_t$. Following the notation in  \cite{sohn2020fixmatch, han2020unsupervised}, $\{x_l, x_l^w\} \in X_l$ are the paired labeled samples with labels $y_l \in \{1, \dots, C\}$, and $\{x_u, x_u^w, x_u^s\} \in X_u$ are the triple unlabeled samples. Here, $x_l$ and $x_u$ represent the raw images without any transformations. ${x_l^w, x_u^w}$, and $x_u^s$ are the images based on the weak and strong augmentation strategies, respectively. In this paper, weak augmentation uses a standard flip-and-shift strategy, and strong augmentation uses the RandAugment~\cite{cubuk2020randaugment} strategy with Cutout~\cite{devries2017improved} augmentation operation.

\subsection{Background}
\textbf{FixMatch}. In the FixMatch~\cite{sohn2020fixmatch} algorithm, apart from basic cross-entropy on labeled samples, consistency regularization with pseudo labels on unlabeled samples is represented as:
\begin{equation}
\label{eq:fixmatch}
\mathcal{L}^{FM}(x_u^w, x_u^s) = \mathbbm{1}(max(p(y|x_u^w)) \ge \epsilon) \cdot H(\hat{y}_u^w, p(y|x_u^s)),
\end{equation}
where $H$ is the cross-entropy, $\tau$ is the pre-defined threshold, $p$ represents the output probability of the model, and $\hat{y}_u^w:=\arg max (p(y|x_u^w))$ is the pseudo label from the weakly augmented predictions.

\textbf{Multiple Choice Learning}. Stochastic Multiple Choice Learning~(sMCL)~\cite{lee2016stochastic} aims to specialize each individual model on a subset of data, by minimizing the loss as follows:
\begin{equation}
\label{eq:smcl}
   \mathcal{L}^{sMCL}{(x_l, y_l)} = \underset{k \in \{1,\cdots,K\}}{\min}{ H(y_l, p_k(y|x_l))},
\end{equation}
where $p_k$ is the output probability of the $k$-th model. For a training sample $(x_l, y_l)$, sMCL feeds the data to all $K$ models but only chooses the most accurate model to do back-propagation. Consequently, each model performs better on some classes than the other models, \emph{i.e.}, each model becomes a specialist on some particular classes.

\subsection{Framework}
\label{sec:framework}
\begin{figure*}[t]
	\begin{center}
		\includegraphics[width=\linewidth]{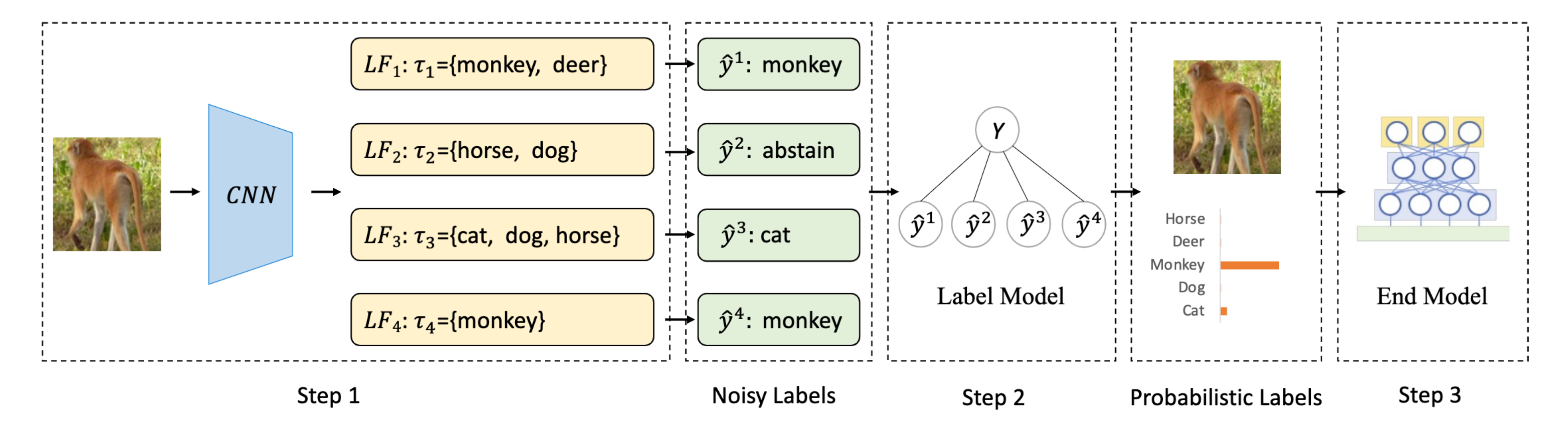}
	\end{center}
	\caption{Framework of the DP-SSL method with four LFs.}
	\label{fig:framework}
\end{figure*}

Fig.~\ref{fig:framework} shows the framework of our DP-SSL method, which works in three major steps as follows:
\begin{itemize}
	\item Step 1. We employ an MCL based approach to automatically generate $K$ LFs from scratch in an SSL style. Here, each LF is trained on a subset of $C$ classes in  the training set based on MCL. As shown in Fig~\ref{fig:framework}, the 2nd LF is trained with samples of classes ``horse'' and ``dog'', and abstains from predicting when facing monkey images. 
	\item Step 2.
	A graphical model is developed as the label model to aggregate the noisy labels and produce probabilistic labels for unlabeled training data.
	The label model is learned in an SSL manner with an additional regularizer.
	\item Step 3. The end model is trained with both provided labels and probabilistic labels generated from Step 2. Finally, we verify the performance of the end model on the test data.
\end{itemize}

\subsection{Labeling Function}
\label{sec:method-LF}
In Step 1 of our method, LFs are exploited to generate noisy labels for each unlabeled image. In previous DP works for computer vision, LFs are built via external image-agnostic knowledge~\cite{chen2019scene} or pretrained models~\cite{Hooper2021cutout, pal2018adversarial,das2020goggles}.
However, 
it is difficult to explicitly describe the rules of image classification. Instead, here we innovatively explore MCL and SSL for automatic LF generation.

As shown in Fig.~\ref{fig:framework}, we share the same backbone~(Wide ResNet~\cite{zagoruyko2016wide} in this paper) to extract features of images for multiple prediction heads~(called LFs in this paper).
To promote the diversity of LFs, we transform the features and feed each LF with different transformed features as follows: 
\begin{equation}
	\label{eq:vlad}
	{f}_{k} = \sum_{j=1}^{HW}{\frac{e^{-\beta_k dis(f[j], c_k)}}{\sum_{k'=1}^K e^{-\beta_{k'} dis(f[j], c_{k'})}}} (f[j] - c_k).
\end{equation}
In this paper, $K$ is the number of LFs, $f \in \mathbb{R}^{HW \times D}$ denotes the feature map of input image $x$ before global average pooling of backbone, $f[j] \in \mathbb{R}^{D} $ is the feature vector at the spatial position $j$ of $f$. $c_k \in \mathbb{R}^{D}$ is the learnable clustering center of the $k$-th LF, 
$\beta_k$ is the learnable variable of the $k$-th cluster,
$dis(A, B)$ represents the distance between $A$ and $B$.
Thus, ${f}_{k}$ corresponds to the feature fed to the $k$-th LF, and describes the $k$-th aggregated pattern of $f$ among the $K$ centers, it can also be considered as a learnable weighted spatial pooling for feature $f$. 
Then, supposing $\mathcal{F}_k$ is the classifier in the $k$-th LF, which would output the probability $\mathcal{F}_k(f_k)$ as prediction.
For clarity, in the following we denote $p_k(y|x):=\mathcal{F}_k(f_k)$ of the $k$-th LF with the input image $x$.

As depicted in \cite{lee2016stochastic}, the classifiers lack diversity of prediction even trained with different protocols. Therefore, we adopt MCL to assign a subset of labeled data for each classifier automatically to improve diversity. However, it is intuitive to observe that in Eq.~(\ref{eq:smcl}) each category can only be assigned to one LF, and no consensus can be exploited.
Therefore, we increase the proportion of selected models in Eq.~(\ref{eq:smcl}) to do back-propagation, which is formulated as
\begin{equation}
	\label{eq:mcl}
	\mathcal{L}_l^{MCL}(x_l^w, y_l) = \underset{\begin{subarray}{c}
			\mathcal{M}\subset \{H(y_l, p_k(y|x_l^w)\}_{k=1}^{K}  \\
			|\mathcal{M}|= \rho\cdot K
		\end{subarray}}{\min}  \frac{1}{ \rho\cdot K} \sum_{k'=1}^{ \rho\cdot K} \mathcal{M}_{k'},
\end{equation}
where $\mathcal{M}_{k'}$ indicates the $k'$-th element in the set $\mathcal{M}$, and $\rho \in [1/K, 1]$ is a designed parameter to represent the ratio of specialist LFs. When $\lfloor\rho\cdot K\rfloor$ is equal to 1, Eq.~(\ref{eq:mcl}) becomes the traditional MCL in Eq.~(\ref{eq:smcl}). In contrast, if $\lfloor\rho\cdot K\rfloor$ is equal to K, it deteriorates to the basic ensemble learning, where all $K$ classifiers are trained with the same data.

Based on MCL, each LF is a specialist for some classes, so it can get high accuracy for samples in these classes. While for samples from other classes not specialized by the LF,  it fails to predict due to over-confidence. Thus, we take only the probabilities of specialized categories as predictions, and allow each LF to abstain from some samples in the dataset. Formally, we denote `0' as the abstention label, and the specialized category set of the $k$-th LF as $\tau_k = \{\tau_k^1,  \dots, \tau_k^{| \tau_k|}\}$. Then, the output label of the $k$-th LF $\hat{y}^k$ satisfies $\hat{y}^k \in \{0\}\cup \tau_k$, \emph{e.g.}, the output of the $1$st LF in Fig.~\ref{fig:framework} is among ``monkey'', ``deer'' and ``abstention'' because its specialized category set $\tau_1=\{$monkey, deer$\}$. 
Then, we denote the probability over the specialized category set $\tau_k$ and ``abstention'' option as $\bar{p}_{k}(y|x)$, where $\bar{p}_k(y|x)\in\mathbb{R}^{|\tau_k| + 1}$.
The objective function over labeled samples with abstention option is
\begin{equation}
	\label{eq:mcl-l2}
	\mathcal{L}_l(x_l^w, y_l) = \sum_{k=1}^K{\big(\mathbbm{1}(y_l \in \tau_k)H(y_l, \bar{p}_k(y|x_l^w)) +
		\mathbbm{1}(y_l \notin \tau_k)H(0, \bar{p}_k(y|x_l^w))\big)},
\end{equation} 
Then, for the unlabeled training data, we follow the settings in FixMatch~\cite{sohn2020fixmatch}, where unlabeled data are supervised by the pseudo labels $\hat{y}_u^{w,k}$ of weak augmentation data $x_u^w$. Thus,
\begin{equation}
	\label{eq:ssl}
	\begin{split}
	\mathcal{L}_u(x_u^w, x_u^s) = \sum_{k=1}^{K} \mathbbm{1}(\max(\bar{p}_k(y|x_u^w)) \ge \epsilon)   \Big( \mathbbm{1}(\hat{y}_u^{w,k} \in \tau_k ) H(\hat{y}_u^{w,k}, \bar{p}_k(y|x_u^s)) + \\  \mathbbm{1}(\hat{y}_u^{w,k} \notin \tau_k)H(0, \bar{p}_k(y|x_u^w)) \Big),    \\
	\end{split}
\end{equation}
where $\hat{y}_u^{w,k} := arg\max(\bar{p}_k(y|x_u^w))$. Specifically, we only keep samples whose largest probability (including the abstention option) is above the predefined threshold $\epsilon$~(0.95 in our paper), and train the model on the kept data with pseudo label $\hat{y}_u^{w,k}$. Accordingly, the training in this step is to minimize the objective function as follows:
\begin{equation}
	\label{eq:loss}
	\mathcal{L}(x_l^w, y_l, x_u^w, x_u^s) = \mu_l^{MCL}\mathcal{L}_l^{MCL}(x_l^w, y_l) + \mu_l \mathcal{L}_l(x_l^w, y_l) + \mu_u \mathcal{L}_u(x_u^w, x_u^s),
\end{equation}
where $\mu_l^{MCL}$, $\mu_{l}$ and $\mu_u$ are hyper-parameters. In our implementation, we first set $\mu_l^{MCL}=1$ and $\mu_l=\mu_u=0$, then adjust $\mu_l^{MCL}$ to $0$ and $\mu_l=\mu_u=1$ after the convergence of $\mathcal{L}_l^{MCL}$.

Generally, in Step 1, MCL is expected to generate specialized class sets $\tau$ for LFs, with which samples are more easily discriminated by SSL classifiers even there are a few labeled samples. Besides, the abstention option is for addressing the over-confidence problem of samples from non-specialized sets.

\subsection{Label Model}
\label{sec:pgm}
In Step 2 of our method, we utilize a graphical model to specify a single prediction by integrating noisy labels provided by $K$ LFs. For simplification, we assume that the $K$ LFs are independent~({as shown in Fig.~\ref{fig:framework}}). 
Then, suppose that $\hat{\mathbf{y}} = (\hat{y}^1, \cdots, \hat{y}^K)^\intercal \in \mathbb{R}^{K}$ is the vectorized form of the predictions from $K$ LFs, the joint distribution of the label model can be described as:
\begin{equation}
	\label{eq:distribution}
	P(y, \hat{\mathbf{y}}) = \frac{1}{Z}\prod_{k=1}^{K}\phi(y, \hat{y}^k)
\end{equation}
where $Z$ is the normalizer of the joint distribution, $\phi$ is the potential that couples the target $y$ and noisy label $\hat{y}^k$. 
In this paper, we extend the dimension of parameters $\theta$ in label model to $K \times C$ to support multi-class classification. 
Set $e_{ky}:=exp(\theta_{ky})$, which is the exponent of parameters $\theta_{ky}$.
Now we are to construct the potential function $\phi$.
Due to the specialized LFs, the potential $\phi$ should benefit the final prediction when a noisy label agrees with the target. That is, we should have $\phi(y, \hat{y}^k) > 1$. Thus, we set $\phi$ as $1+e_{ky}$ for this case.
On the contrary, the potential $\phi$ should negatively impact the final prediction when a noisy label conflicts with the target label in the specialized category set, i.e., we should have $\phi(y, \hat{y}^k) < 1$. Therefore, for this case we set $\phi$ to $1/(1+e_{ky})$. 
For the other cases, we follow the design in \cite{ratner2020snorkel}. 
In summary, %we abbreviate $exp(\theta_{ky})$ as $e_{ky}$,
the potential $\phi$ is defined as follows:
\begin{equation}
	\label{eq:potential}
	\phi(y,\hat{y}^k)=\left\{\begin{aligned}
		&1 + e_{ky}, &{ if \; y \in \tau_k,\;\hat{y}^k\in \tau_k,\; \hat{y}^k=y}\\
		&1/(1+e_{ky}), &if \; y \in \tau_k,\;\hat{y}^k\in \tau_k,\; \hat{y}^k \neq y\\
		&e_{ky}, &if \; y \notin \tau_k, \; \hat{y}^k \in \tau_k,\; \hat{y}^k \neq y \\
		&1. &otherwise
	\end{aligned}\right.
\end{equation}
With the potential above, the normalizer $Z$ of the joint distribution in Eq.~(\ref{eq:distribution}) can be obtained by summarizing over $y$ and $\hat{y}^k$:
\begin{equation}
	\begin{split}
		Z&=\sum_{y \in \mathcal{Y}}\prod_{k=1}^{K}\sum_{\hat{y}^k \in \{0\}\cup\tau_k} \phi(y, \hat{y}^k) \\
		&= \sum_{y \in \mathcal{Y}}\prod_{k=1}^{K} \Big(\mathbbm{1}(y \in \tau_k)({2}+e_{ky}+\frac{|\tau_k| -1}{1 + e_{ky}}) +  \mathbbm{1}(y \notin \tau_k)(1+|\tau_k|e_{ky}) \Big).
	\end{split}
\end{equation}
Then, the objective function of the label model can be expressed in an SSL manner as follows:
\begin{equation}
	\label{eq:loss-pgm}
	\mathcal{L}(\hat{\mathbf{y}}_l, y_l, \hat{\mathbf{y}}_u) =
	\underbrace{\sum_{x_l}H(y_l, P(y, \hat{\mathbf{y}}_l))
	}_{\text{labeled samples}} +
	\underbrace{(-\sum_{x_u} \log \sum_{y \in \mathcal{Y}}P(y, \hat{\mathbf{y}}_u))}_{\text{unlabeled samples}} + R(\theta, \hat{\mathbf{y}}_u),
\end{equation}
where the first part is the cross-entropy loss, the second is the negative log marginal likelihood on the observed noisy labels $\hat{\mathbf{y}}_u$, and the third is a regularizer.
In our method, the regularizer is utilized to guide the label model with statistical information (the accuracy of each LF).
%However, {\color{red}the accuracy of LF on labeled training data is unreliable to transfer to the unlabeled data}.
However, the accuracy of each LF on noisy labels is unavailable, while the accuracy on labeled training is almost 100\% due to over-fitting.
Thus, we have to estimate the accuracy of each LF with the observable noisy labels $\hat{\mathbf{y}}$, which will be presented in Sec.~\ref{sec:acc}.
After training, the label model produces probabilistic labels $\pi$ by computing the joint distribution in Eq.~(\ref{eq:distribution}) with the noisy labels $\hat{\mathbf{y}}$.

\subsection{Accuracy Estimation}
\label{sec:acc}
Now, we formally describe our method for estimating the accuracy of LFs. We transform the multi-class problem into $C$ one-versus-all tasks. For the $i$-th ($i \in [1,\cdots, C]$) one-versus-all task, we denote the unobserved ground-truth labels as $z_i\in\{\pm1\}$~($z_i=+1$ means $y=i$, and $z_i=-1$ represents $y\neq i$), noisy labels of the $k$-th LF as $\hat{z}_i^k\in\{\pm1,0\}$,
\begin{equation}
	\hat{z}_i^k= \left\{
	\begin{matrix}
		1 & if  \; \hat{y}^k=i, \\
		0 & if \; \hat{y}^k=0, \\
		-1 & otherwise.
	\end{matrix}\right.
\end{equation}
Then, we can write $\mathbb{E}[\hat{z}_i^kz_i]$ as
\begin{equation}
	\label{eq:E}
	\begin{split}
		\mathbb{E}[\hat{z}_i^kz_i] &= \mathit{P}(\hat{z}_i^kz_i=1) -  \mathit{P}(\hat{z}_i^kz_i=-1)\\
		&= \mathit{P}(\hat{z}_i^kz_i=1) - (1 - \mathit{P}(\hat{z}_i^kz_i=1) - \mathit{P}(\hat{z}_i^kz_i=0))\\
		&=2\mathit{P}(\hat{z}_i^k=z_i) + \mathit{P}(\hat{z}_i^k=0)-1.
	\end{split}
\end{equation}
Assume that $\hat{z}_i^{j} \independent \hat{z}_i^{k} ~|z_i$ for distinct ${j}$ and ${k}$, then
\begin{equation}
	\label{eq:EE}
	\mathbb{E}[\hat{z}_i^{j}\hat{z}_i^{k}] = \mathbb{E}[\hat{z}_i^{j}z_i^2\hat{z}_i^{k}] = \mathbb{E}[\hat{z}_i^{j}z_i] \mathbb{E}[\hat{z}_i^{k}z_i]
\end{equation}
with the fact that $z_i^2=1$. In Eq.~(\ref{eq:EE}),  $\hat{\mathbb{E}}[\hat{z}_i^{j}\hat{z}_i^{k}] =\frac{1}{|x_u|}\sum_{x_u}\hat{z}_i^j\hat{z}_i^k $ is observable, which can be derived from the noisy labels of the $j$-th and $k$-th LFs, while $\mathbb{E}[\hat{z}_i^{j}z_i]$ and $\mathbb{E}[\hat{z}_i^{k}z_i]$ remain to be solved due to true label $z_i$ is unavailable. Next, we introduce a third labeling result from the $l$-th LF as $\hat{z}_i^l$, such that $\hat{\mathbb{E}}[\hat{z}_i^{j}\hat{z}_i^{l}]$ and $\hat{\mathbb{E}}[\hat{z}_i^{k}\hat{z}_i^{l}]$ are observable. Then, $|\hat{\mathbb{E}}[\hat{z}_i^{j}z_i]|$, $|\hat{\mathbb{E}}[\hat{z}_i^{k}z_i]|$, $|\hat{\mathbb{E}}[\hat{z}_i^{l}z_i]|$ can be solved by a triplet method as follows:
\begin{equation}
	\label{eq:triplet}
	\begin{split}
		|\hat{\mathbb{E}}[\hat{z}_i^{j}z_i]| &= \sqrt{|\hat{\mathbb{E}}[\hat{z}_i^{j}\hat{z}_i^{k}]\cdot \hat{\mathbb{E}}[\hat{z}_i^{j}\hat{z}_i^{l}]/\hat{\mathbb{E}}[\hat{z}_i^{k}\hat{z}_i^{l}]|}, \\
		|\hat{\mathbb{E}}[\hat{z}_i^{k}z_i]| &= \sqrt{|\hat{\mathbb{E}}[\hat{z}_i^{j}\hat{z}_i^{k}]\cdot \hat{\mathbb{E}}[\hat{z}_i^{k}\hat{z}_i^{l}]/\hat{\mathbb{E}}[\hat{z}_i^{j}\hat{z}_i^{l}]|}, \\
		|\hat{\mathbb{E}}[\hat{z}_i^{l}z_i]| &= \sqrt{|\hat{\mathbb{E}}[\hat{z}_i^{j}\hat{z}_i^{l}]\cdot \hat{\mathbb{E}}[\hat{z}_i^{k}\hat{z}_i^{l}]/\hat{\mathbb{E}}[\hat{z}_i^{j}\hat{z}_i^{k}]|}. \\
	\end{split}
\end{equation}
We can obtain the estimated accuracy of each LF by resolving the sign of  $\mathbb{E}[\hat{z}_i^kz_i]$~\cite{fu2020fast}.
Let $\hat{a}_i^k:=\hat{\mathit{P}}(\hat{z}_i^k=z_i | \hat{z}_i^k\neq0)$ be the estimated accuracy of the $k$-th LF on the $i$-th category.
Therefore, the regularizer of $R(\theta, \hat{\mathbf{y}}_u)$ can be formulated as
\begin{equation}
	R(\theta, \hat{\mathbf{y}}_u) =\sum_{i=1}^C \sum_k^K \hat{a}_i^k \log \mathit{P}_\theta(\hat{z}_i^k=z_i|\hat{z}_i^k\neq 0) + (1 - \hat{a}_i^k)\log (1 - \mathit{P}_\theta(\hat{z}_i^k=z_i|\hat{z}_i^k\neq 0))
\end{equation}
where $\mathit{P}_\theta(\hat{z}_i^k=z_i|\hat{z}_i^k\neq 0)$ can be computed in closed form by marginalizing over all the other variables in the model in Eq.~(\ref{eq:distribution}). Details of $\mathit{P}_\theta$ can be referred to \textbf{Appendix} \ref{sec:app:reg}.

\subsection{End Model}
In Step 3, probabilistic labels are used to train an end model under any network architecture. We utilize noise-aware empirical risk expectation as the objective function to take annotation errors into account. Accordingly, the final objective function is as follows:
\begin{equation}
	\mathcal{L}(x_l, y_l, x_u, \pi) = \underbrace{\sum_{x_l}H(y_l, p(y|x_l))}_{\text{labeled samples}} +
	\underbrace{\sum_{x_u}\mathbb{E}_{y \sim \pi}H(y, p(y|x_u))}_{\text{unlabeled samples with probabilistic label}}
\end{equation}
where $p(y|x_l)$ and $p(y|x_u)$ are the predicted distributions of $x_l$ and $x_u$, $\pi$ is the distribution produced by the label model in Sec.~\ref{sec:pgm}.  

\section{Experiments}
\subsection{Implementation Details}
\label{sec:exp-detail}
In the training phase, we follow the settings of previous works~\cite{sohn2020fixmatch, berthelot2019remixmatch, han2020unsupervised}, augment data in weak~(a standard flip-and-shift strategy) and strong forms~(RandAugment~\cite{cubuk2020randaugment} followed by Cutout~\cite{devries2017improved} operation), and utilize a Wide ResNet as the end model for a fair comparison. In our framework, the batch size for labeled data and unlabeled data is set to 64 and 448, respectively. Besides, we use the same hyperparameters~($K=50$, $\rho=0.2$, $\epsilon=0.95$) for all datasets.
We compare DP-SSL with major existing methods on CIFAR-10~\cite{krizhevsky2009learning}, CIFAR-100~\cite{krizhevsky2009learning}, SVHN~\cite{netzer2011reading} and STL-10~\cite{coates2011analysis}.
We also analyze the effect of annotation and conduct ablation study in Sec.~\ref{sec:annotation} and Sec.~\ref{sec:ablation} respectively. All experiments are implemented in Pytorch v1.7 and conducted on 16 NVIDIA RTX3090s.

\subsection{Datasets}
\label{sec:dataset}
\textbf{CIFAR-10 and CIFAR-100} \cite{krizhevsky2009learning} contain 50,000 training examples and 10,000 validation examples. All images are of 32x32 pixel size and fall in 10 or 100 classes, respectively.

\textbf{SVHN} \cite{netzer2011reading} is a digital image dataset that consists of 73,257,  26,032 and 531,131 samples in the train, test, and extra folders. It has the same image resolution and category number as CIFAR-10.

\textbf{STL-10} \cite{coates2011analysis} is a dataset for evaluating unsupervised and semi-supervised learning. It consists of 5000 labeled images and 8000 validation samples of 96x96 size from 10 classes. Besides, there are 100,000 unlabeled images available, including odd samples.

\subsection{Comparison with Existing SSL Methods}
\label{sec:full-exp}
For a fair comparison, we conduct experiments with the codebase of FixMatch and cite the results on CIFAR-10, CIFAR-100, SVHN and STL-10 from \cite{sohn2020fixmatch, han2020unsupervised}. We utilize the same network architecture (a Wide ResNet-28-2 for CIFAR-10 and SVHN, WRN-28-8 for CIFAR-100, and WRN-37-2 for STL-10) and training protocol of FixMatch, such as optimizer and learning rate schedule. Unlabeled data are generated by the scripts in FixMatch. Results of DP-SSL and existing methods in Tab.~\ref{tab:sota1} and Tab.~\ref{tab:sota-STL10} are presented with the mean and standard deviation~(STD) of accuracy on 5 pre-defined folds.

As shown in Tab.~\ref{tab:sota1}, our method achieves the best performance in most cases, especially when there are only 4 labeled samples per class. Specifically, our method achieves a 93.46\% accuracy on CIFAR-10 with 4 labeled samples per category, which is 3.3\% higher than that of USADTM --- the state-of-the-art method. Again on STL-10, our method surpasses USADTM and achieves the best performance when there are 4 and 25 labeled samples per class.

On CIFAR-100, our method performs the best for 400 labels case and the 2nd for 2500 and 10,000 labels cases. We also notice that DP-SSL has relatively large STDs for 2500 and 10,000 labels cases, which is due to the coarse accuracy estimation. In fact, even if triplet mean is adopted in estimation, the triplet selection in Eq.~(\ref{eq:triplet}) still impacts accuracy estimation and regularizer a lot, especially when $\mathbb{E}[\hat{z}_i^kz_i]$ is close to 0 or sign recovery of $\mathbb{E}[\hat{z}_i^kz_i]$ is wrong.
Actually, there are some advanced approaches to unsupervised accuracy estimation~\cite{jaffe2015estimating,platanios2017estimating, traganitis2018blind} that can replace the naive triplet mean estimation.
Ideally, if we can obtain the exact accuracy of each class $\hat{b}_i^k:=\hat{\mathit{P}}(\hat{z}_i^k=z_i | \hat{z}_i^k=1)$ and regularize it as $R(\theta, \hat{\mathbf{y}}_u) =\sum_{i=1}^C \sum_k^K \hat{b}_i^k \log \mathit{P}_\theta(\hat{z}_i^k=z_i|\hat{z}_i^k=1) + (1 - \hat{b}_i^k)\log (1 - \mathit{P}_\theta(\hat{z}_i^k=z_i|\hat{z}_i^k=1))$, we will get an end model with ($27.92\pm\textbf{0.23}$)\% error rate for 2500 labeled samples.

Comparing with USADTM, our method does not perform well enough when more labeled data available. For USADTM, apart from the proxy label generator, unsupervised representation learning contributes a lot for its performance.
As shown in the ablation study of \cite{han2020unsupervised}, USADTM without unsupervised representation learning  achieves around 5.73\% and 4.99\% error rate for 250 and 4000 labeled samples in CIFAR-10, while our method DP-SSL obtains 4.78\% and 4.23\% error rate.

\begin{table}[t]
	\centering
	\scriptsize
	\centering
	\caption{Results of error rate on CIFAR-10, CIFAR-100 and SVHN for different existing SSL methods ($\Pi$-Model \cite{laine2016temporal}, Pseudo-Labeling \cite{lee2013pseudo}, Mean Teacher \cite{tarvainen2017mean}, MixMatch \cite{berthelot2019mixmatch}, UDA \cite{xie2020unsupervised}, ReMixMatch \cite{berthelot2019remixmatch}, FixMatch \cite{sohn2020fixmatch} and USADTM \cite{han2020unsupervised}) and our DP-SSL method.} \label{tab:sota1}
	\setlength{\tabcolsep}{0.55mm}{
		\begin{tabular}{l|rrr|rrr|rrr}
			\toprule
			&\multicolumn{3}{c|}{CIFAR-10}&\multicolumn{3}{c|}{CIFAR-100}&\multicolumn{3}{c}{SVHN}\\
			\midrule
			Method & 40 labels & 250 labels &4000 labels &400 labels & 2500 labels & 10000 labels  & 40 labels & 250 labels &1000 labels\\
			\midrule
			$\Pi$ -Model                 & -            & 54.26$\pm$3.97   &14.01$\pm$0.38   &-                &57.25$\pm$0.48  &37.88$\pm$0.11 & -            & 18.96$\pm$1.92   &7.54$\pm$0.36   \\
			Pseudo-Labeling              & -            & 49.78$\pm$0.43   &16.09$\pm$0.28   &-                &57.38$\pm$0.46  &36.21$\pm$0.19
			& -            & 20.21$\pm$1.09   &9.94$\pm$0.61  \\
			Mean Teacher                 & -            & 32.32$\pm$2.30   &9.19$\pm$0.19    &-                &53.91$\pm$0.57  &35.83$\pm$0.24
			& -            & 3.57$\pm$0.11   &3.42$\pm$0.07    \\
			MixMatch                  & 47.54$\pm$11.50 & 11.05$\pm$0.86   &6.42$\pm$0.10    & 67.61$\pm$1.32  &39.94$\pm$0.37  &28.31$\pm$0.33
			& 42.55$\pm$14.53 & 3.98$\pm$0.23   &3.50$\pm$0.28\\
			UDA                       & 29.05$\pm$5.93  & 8.82$\pm$1.08    &4.88$\pm$0.18    & 59.28$\pm$0.88  &33.13$\pm$0.22  &24.50$\pm$0.25
			& 52.63$\pm$20.51 & 5.69$\pm$2.76    &2.46$\pm$0.24\\
			ReMixMatch                & 19.10$\pm$9.64  & 5.44$\pm$0.05    &4.72$\pm$0.13    & 44.28$\pm$2.06  &\textbf{27.43}$\pm$0.31  &23.03$\pm$0.56
			& 3.34$\pm$0.20  & 2.92$\pm$0.48    &2.65$\pm$0.08  \\
			USADTM  & 9.54$\pm$1.04 & 4.80$\pm$0.32 &4.40$\pm$0.15  & 43.36$\pm$1.89  &28.11$\pm$0.21  &\textbf{21.35}$\pm$0.17 & 3.01$\pm$1.97  & \textbf{2.11}$\pm$0.65   &\textbf{1.96}$\pm$0.05  \\
			\toprule
			FixMatch (RA)             & 13.81$\pm$3.37  & 5.07$\pm$0.65    &4.26$\pm$0.05  & 48.85$\pm$1.75  &28.29$\pm$0.11  &22.60$\pm$0.12
			& 3.96$\pm$2.17  & 2.48$\pm$0.38    &2.28$\pm$0.11  \\
			FixMatch (CTA)            & 11.39$\pm$3.35  & 5.07$\pm$0.33    &4.31$\pm$0.15    & 49.95$\pm$3.01  &28.64$\pm$0.24  &23.18$\pm$0.11
			& 7.65$\pm$7.65  & 2.64$\pm$0.64    &2.36$\pm$0.19 \\
			DP-SSL (ours) & \textbf{6.54}$\pm$0.98  & \textbf{4.78}$\pm$0.26  & \textbf{4.23}$\pm$0.20 & \textbf{43.17}$\pm$1.29 & 28.00$\pm$0.79 &  22.24$\pm$0.31 & \textbf{2.98}$\pm$0.86 & 2.16$\pm$0.36 & 1.99$\pm$0.18 \\
			\midrule
			Fully Supervised          & \multicolumn{3}{c|}{2.74}&\multicolumn{3}{c}{16.84} & \multicolumn{3}{|c}{1.48}\\
			\bottomrule
	\end{tabular}}
\end{table}
\begin{table}[t]
	\centering
	\scriptsize
	\caption{Results of error rate on STL-10.}
	\label{tab:sota-STL10}
	\centering
	\setlength{\tabcolsep}{1.8mm}{
		\begin{tabular}{l|r|l|r|l|rrr}
			\toprule
			\multicolumn{8}{c}{STL-10}\\
			\midrule
			Method  &1000 labels &Method & 1000 labels &Method &40 labels & 250 labels & 1000 labels \\
			\midrule
			$\Pi$ -Model     &26.23$\pm$0.82 &UDA            &7.66$\pm$0.56  &USADTM   & 9.63$\pm$1.35 &6.85$\pm$1.09  &\textbf{4.01}$\pm$0.59   \\
			Pseudo-Labeling  &27.99$\pm$0.80 &ReMixMatch    &5.23$\pm$0.45  & DP-SSL (ours)  & \textbf{9.32}$\pm$0.91 &\textbf{6.83}$\pm$0.71 & 4.97$\pm$0.42 \\
			Mean Teacher     &21.43$\pm$2.39 &FixMatch (RA)  &7.98$\pm$1.50  & Fully Supervised & \multicolumn{3}{c}{1.48}\\
			MixMatch         &10.41$\pm$0.61 &FixMatch (CTA) &5.17$\pm$0.63  \\
			\bottomrule
	\end{tabular}}
\end{table}

\subsection{Analysis}
\label{sec:annotation}
\textbf{Annotation performance}.
Intuitively, the holistic performance of the end model in our method highly depends on the quality of annotation results. Thus, we present the macro precision/recall/F1 score and coverage of the annotated labels of our method on CIFAR-10, CIFAR-100, and SVHN in Tab.~\ref{tab:label-res}. We can see that our method achieves over 99\% coverage, which means that it produces probabilistic labels for almost all unlabeled data. Comparing to the results in \cite{han2020unsupervised}, the label model with 40 labeled samples outperforms the proxy label generator, FixMatch and USADTM get 88.51\% and 89.48\% accuracy, respectively. Furthermore, our method achieves 97.36\% accuracy for unlabeled data with the top-500 highest probabilities in each category. Meanwhile, we also present results of Majority Voting and FlyingSquid~\cite{fu2020fast} in Tab.~\ref{tab:label-res} based on the noisy labels from Step 1 of our method for comparison. Majority Voting gets bad performance because the number of LFs triggered for different categories is not equal. For FlyingSquid, we implement it with $C$ one-versus-all models to support multi-class tasks, and the large $C$ in CIFAR-100 results in the worst performance.

\textbf{Barely supervised learning}.
We conduct experiments to test the performance~(accuracy and STD) of our method on CIFAR-10 for some extreme cases~(10, 20 and 30 labeled samples) to verify the effectiveness of our method.
Here, we select the labeled data through the scripts of FixMatch with 5 different random seeds.
As claimed in FixMatch, it reaches between 48.58\% and 85.32\% test accuracy with a median of 64.28\% for 10 labeled samples, while our method obtains accuracy from 61.32\% to 83.7\%.
As for 20 and 30 labeled samples, our method gets ($85.29\pm3.14$)\% and ($89.81\pm1.59$)\% accuracy respectively, which have much smaller STDs than that reported in~\cite{li2020comatch}.

\begin{table}
	\centering
	\scriptsize
	\centering
	\caption{The macro Precision/Recall/F1 Score/Coverage of the annotated labels on CIFAR-10, CIFAR-100, and SVHN for our method and two typical existing label models.}
	\label{tab:label-res}
	\setlength{\tabcolsep}{0.65mm}{
		\begin{tabular}{l|l|ccc|ccc|ccc}
			\toprule
			\multicolumn{2}{c|}{ }	&\multicolumn{3}{c|}{CIFAR-10}&\multicolumn{3}{c|}{CIFAR-100}&\multicolumn{3}{c}{SVHN}\\
			\midrule
			Method & Metrics & 40 labels & 250 labels &4000 labels &400 labels & 2500 labels & 10000 labels  & 40 labels & 250 labels &1000 labels\\
			\midrule
			Majority Vote                             & F1 Score & 85.96 & 94.23 & 95.77 & 49.97  & 69.81  & 76.03& 90.86 & 95.38 & 96.14\\
			FlyingSquid\cite{fu2020fast} & F1 Score & 90.25 & 94.99 & 95.85 & 48.90  & 69.73 & 74.12 & 93.92 & 97.24 & 97.70 \\
			\midrule
			\multirow{4}{*}{DP-SSL (ours)} &
			Precision & 93.47 & 95.30  & 95.89 & 55.62  & 71.91 & 75.12 &  95.20 & 97.65 & 97.79 \\
			& Recall &93.82 & 95.33 & 95.91 & 56.86& 72.01 & 78.35 & 96.78 &  97.64 & 97.94 \\
			& F1 Score & \textbf{93.61} & \textbf{95.19} & \textbf{95.90} & \textbf{54.42} & \textbf{71.89} & \textbf{76.36} & \textbf{95.95} & \textbf{97.59} & \textbf{97.81} \\
			& Coverage & 99.35 &  99.79 & 99.91 & 99.33 & 99.87 & 99.94 & 99.15 & 99.67 & 99.93\\
			\bottomrule
	\end{tabular}}
\end{table}

\subsection{Ablation Study}
\label{sec:ablation}

\begin{wraptable}{r}{0.39\linewidth}
	\centering
	\scriptsize
	\caption{Annotation performance for different configurations on CIFAR-10 with 40 and 250 labels. $K$ and $\rho$ are set to 50 and 0.2 by default.}
	\setlength{\tabcolsep}{0.7mm}{
		\begin{tabular}{lcc}
			\toprule
			Experiments & 40 labels & 250 labels\\
			\toprule
			Exp1: \emph{w.o.} MCL & 92.46 & 95.02 \\
			Exp2: MCL \emph{w.o.} FT & 91.61 & 94.98  \\
			Exp3: MCL \emph{w.} FT	& \textbf{93.82} & \textbf{95.33} \\
			\toprule
			Exp4: \emph{w.o.} Regularizer  & 93.19 & 94.94 \\
			Exp5: Regularizer & \textbf{93.82} & \textbf{95.33}  \\
			\bottomrule
	\end{tabular}}
	\label{tab:ablation}
\end{wraptable}
In DP-SSL, LFs and the label model are the core components to assign probabilistic labels for training the end model. Here, we check the effects of the following factors in the process of producing probabilistic labels by taking CIFAR-10 as the example.  For ease of exposition, only the accuracy of predicted labels is presented in Tab.~\ref{tab:ablation}.

\textbf{MCL}. Feature transformation~(FT) described in Eq.~(\ref{eq:vlad}) can be regarded as a weighted spatial pooling for extracted features. It is proposed to boost the diversity of generated LFs.
We conduct comparative experiments for three configurations: 1) \emph{Exp1}: \emph{w.o.} MCL, 2)~\emph{Exp2}: MCL \emph{w.o.} FT, 3)~\emph{Exp3}: MCL \emph{w.} FT. The results are presented in Tab.~\ref{tab:ablation}.
It is interesting to see that \emph{Exp1} is better than \emph{Exp2} but worse than \emph{Exp3}. In fact, \emph{Exp1} is a simple ensemble model with a shared backbone, where each LF is trained independently and predicts the labels within $C$ categories.
In \emph{Exp2}, we observe that some classifiers have never been optimized in the training phase and thus have an empty specialized set when only a few labeled samples per class are available. Moreover, the specialized sets of many LFs are duplicate, which incurs a negative impact on the performance.
However, MCL with FT addresses the drawbacks and helpes our method obtain versatile LFs.

\begin{figure}
	\label{fig:ab}
	\begin{tabular}{c c}
	  \begin{subfigure}[b]{0.46\textwidth}
	\centering
	\includegraphics[width=\textwidth]{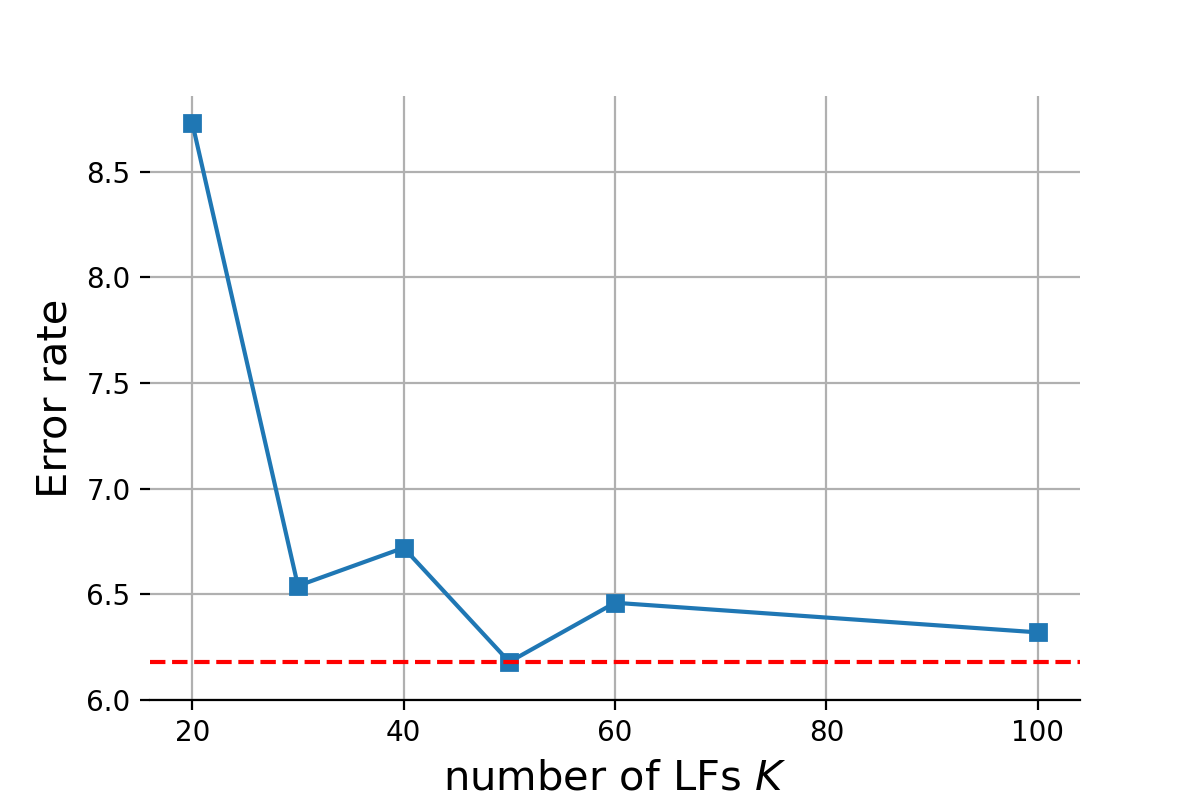}
	\subcaption{}
	\label{fig:ab-a}
	\end{subfigure}   &
	\begin{subfigure}[b]{0.46\textwidth}
	\includegraphics[width=\textwidth]{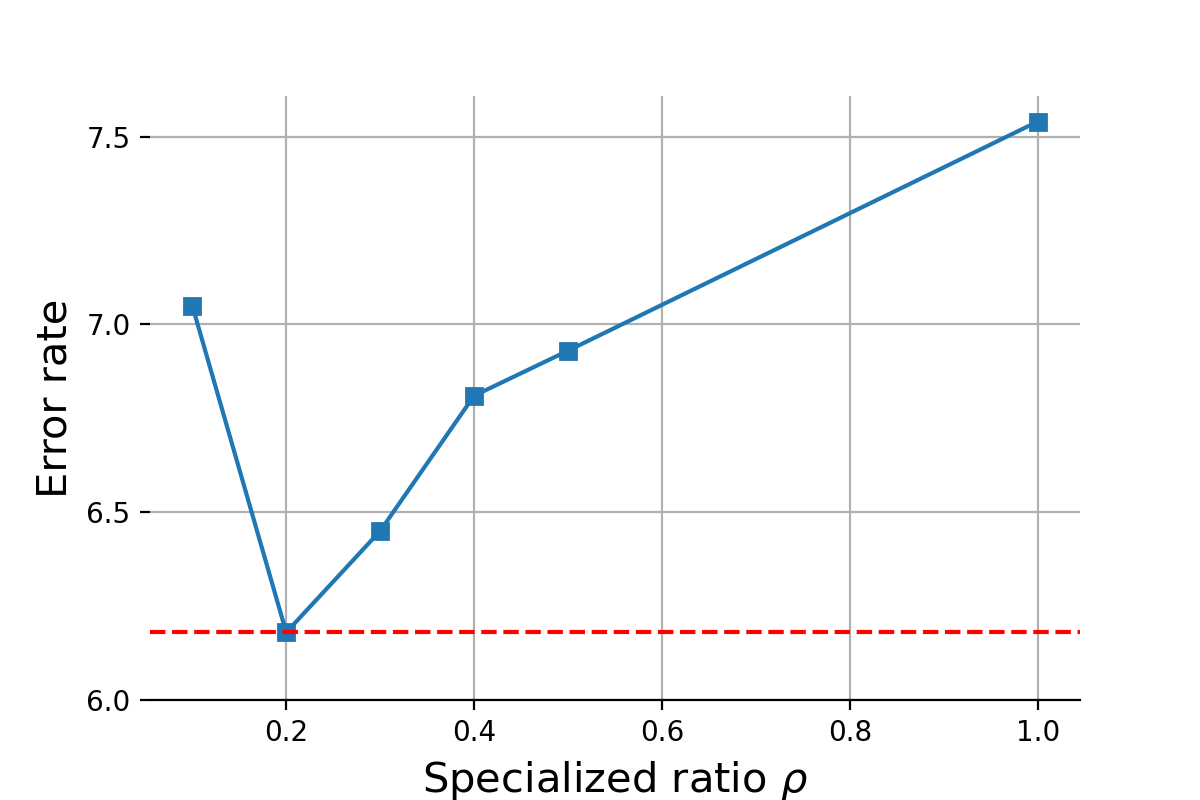}
	\subcaption{}
	\label{fig:ab-b}
	\end{subfigure}
	\end{tabular}
	\caption{Plots of ablation study on CIFAR-10 with 40 labels. (a) Varying the number of LFs K. (b) Varying the specialized ratio $\rho$. Here, the red dashed line indicates the error rate of DP-SSL with default hyperparameters.}
\end{figure}

\textbf{Hyperparameters}. $K$ and $\rho$ are the number of LFs and the ratio of specialists in Eq.~(\ref{eq:mcl}). In our ablation study,
we focus on the variance of performance for different $K$ and $\rho$ with 40 labeled samples on CIFAR-10.
In Fig.~\ref{fig:ab-a}, $K$=50 performs the best when 40 labeled samples are available.  
On the other hand, performance reaches the best when $\rho$=0.2 in Fig.~\ref{fig:ab-b}. 
We present more results of $\rho$ and $K$ in Appendix \ref{sec:app:hyper}.

\textbf{Regularizer}. The regularizer is proposed to impose a global guidance and improve the robustness of the label model. As shown in Tab.~\ref{tab:ablation}, the regularizer does boost the accuracy, especially when facing less labeled samples. Besides, as mentioned in Sec.~\ref{sec:full-exp}, the high-quality guidance of the regularizer also reduces the label model's performance variance, thus improves its robustness.

\section{Conclusion}
In this paper, we explore the data programming idea to boost SSL when only a small number of labeled samples available by providing more accurate labels for unlabeled data. To this end, we propose a new SSL method DP-SSL that employs an innovative DP mechanism to automatically generate labeling functions. To make the labeling functions diverse and specialized, a multiple choice learning based approach is developed. Furthermore, we design an effective label model by incorporating a novel potential and a regularizer with estimated accuracy. With this model, probabilistic labels are inferred by resolving the conflict and overlap among noisy labels from the labeling functions. Finally, an end model is trained under the supervision of the probabilistic labels. Extensive experiments show that DP-SSL can produce high-quality probabilistic labels, and outperforms the existing methods to achieve a new SOTA, especially when only a small number of labeled samples are available. 

\section{Limitations of This Work}
\label{sec:limit}
In this work, we use coarse accuracy estimation as the statistic information to guide the label model for simplicity. As described in Sec.~\ref{sec:acc}, we estimate the accuracy $\mathit{P}_\theta(\hat{z}_i^k=z_i|\hat{z}_i^k\neq 0)$, rather than class-wise accuracy $\mathit{P}_\theta(\hat{z}_i^k=z_i|\hat{z}_i^k =1)$.
Besides, we do not consider the dependency between LFs and directly assume they are independent.

\acksection
This work was supported by Alibaba Group through Alibaba Innovative Research Program. Shuigeng Zhou was also partially supported by Science and
Technology Commission of Shanghai Municipality Project
(No.~19511120700), and Shanghai Artificial Intelligence
Innovation and Development Projects funded by Shanghai
Municipal Commission of Economy and Informatization.

{
	\small
	\bibliography{egbib}{}

\begin{thebibliography}{10}
\providecommand{\url}[1]{#1}
\csname url@samestyle\endcsname
\providecommand{\newblock}{\relax}
\providecommand{\bibinfo}[2]{#2}
\providecommand{\BIBentrySTDinterwordspacing}{\spaceskip=0pt\relax}
\providecommand{\BIBentryALTinterwordstretchfactor}{4}
\providecommand{\BIBentryALTinterwordspacing}{\spaceskip=\fontdimen2\font plus
\BIBentryALTinterwordstretchfactor\fontdimen3\font minus
  \fontdimen4\font\relax}
\providecommand{\BIBforeignlanguage}[2]{{%
\expandafter\ifx\csname l@#1\endcsname\relax
\typeout{** WARNING: IEEEtran.bst: No hyphenation pattern has been}%
\typeout{** loaded for the language `#1'. Using the pattern for}%
\typeout{** the default language instead.}%
\else
\language=\csname l@#1\endcsname
\fi
#2}}
\providecommand{\BIBdecl}{\relax}
\BIBdecl

\bibitem{gao2020consistency}
M.~Gao, Z.~Zhang, G.~Yu, S.~{\"O}. Ar{\i}k, L.~S. Davis, and T.~Pfister,
  ``Consistency-based semi-supervised active learning: Towards minimizing
  labeling cost,'' in \emph{ECCV}.\hskip 1em plus 0.5em minus 0.4em\relax
  Springer, 2020, pp. 510--526.

\bibitem{vondrick2013efficiently}
C.~Vondrick, D.~Patterson, and D.~Ramanan, ``Efficiently scaling up
  crowdsourced video annotation,'' \emph{IJCV}, vol. 101, no.~1, pp. 184--204,
  2013.

\bibitem{yao2021visual}
Y.~Yao, A.~Zhang, X.~Han, M.~Li, C.~Weber, Z.~Liu, S.~Wermter, and M.~Sun,
  ``Visual distant supervision for scene graph generation,'' \emph{arXiv
  preprint arXiv:2103.15365}, 2021.

\bibitem{sohn2020fixmatch}
K.~Sohn, D.~Berthelot, N.~Carlini, Z.~Zhang, H.~Zhang, C.~A. Raffel, E.~D.
  Cubuk, A.~Kurakin, and C.-L. Li, ``Fixmatch: Simplifying semi-supervised
  learning with consistency and confidence,'' in \emph{NeurIPS}, vol.~33, 2020.

\bibitem{oquab2015object}
M.~Oquab, L.~Bottou, I.~Laptev, and J.~Sivic, ``Is object localization for
  free?-weakly-supervised learning with convolutional neural networks,'' in
  \emph{CVPR}, 2015, pp. 685--694.

\bibitem{chen2020simple}
T.~Chen, S.~Kornblith, M.~Norouzi, and G.~Hinton, ``A simple framework for
  contrastive learning of visual representations,'' in \emph{ICML}.\hskip 1em
  plus 0.5em minus 0.4em\relax PMLR, 2020, pp. 1597--1607.

\bibitem{lee2013pseudo}
D.-H. Lee \emph{et~al.}, ``Pseudo-label: The simple and efficient
  semi-supervised learning method for deep neural networks,'' in \emph{Workshop
  on ICML}, vol.~3, no.~2, 2013, p. 896.

\bibitem{sajjadi2016regularization}
M.~Sajjadi, M.~Javanmardi, and T.~Tasdizen, ``Regularization with stochastic
  transformations and perturbations for deep semi-supervised learning,'' in
  \emph{NeurIPS}, 2016, pp. 1163--1171.

\bibitem{zoph2020rethinking}
B.~Zoph, G.~Ghiasi, T.-Y. Lin, Y.~Cui, H.~Liu, E.~D. Cubuk, and Q.~Le,
  ``Rethinking pre-training and self-training,'' in \emph{NeurIPS}, vol.~33,
  2020.

\bibitem{ratner2016data}
A.~J. Ratner, C.~De~Sa, S.~Wu, D.~Selsam, and C.~R{\'e}, ``Data programming:
  Creating large training sets, quickly,'' in \emph{NeurIPS}, vol.~29.\hskip
  1em plus 0.5em minus 0.4em\relax NIH Public Access, 2016, pp. 3567--3575.

\bibitem{awasthi2020learning}
A.~Awasthi, S.~Ghosh, R.~Goyal, and S.~Sarawagi, ``Learning from rules
  generalizing labeled exemplars,'' \emph{ICLR}, 2020.

\bibitem{ratner2017snorkel}
A.~J. Ratner, S.~H. Bach, H.~R. Ehrenberg, and C.~R{\'e}, ``Snorkel: Fast
  training set generation for information extraction,'' in \emph{SIGMOD}.\hskip
  1em plus 0.5em minus 0.4em\relax {ACM}, 2017, pp. 1683--1686.

\bibitem{ratner2019training}
A.~Ratner, B.~Hancock, J.~Dunnmon, F.~Sala, S.~Pandey, and C.~R{\'e},
  ``Training complex models with multi-task weak supervision,'' in \emph{AAAI},
  vol.~33, no.~01.\hskip 1em plus 0.5em minus 0.4em\relax {AAAI} Press, 2019,
  pp. 4763--4771.

\bibitem{chatterjee2020data}
O.~Chatterjee, G.~Ramakrishnan, and S.~Sarawagi, ``Data programming using
  continuous and quality-guided labeling functions,'' \emph{arXiv preprint
  arXiv:1911.09860}, 2019.

\bibitem{chen2019scene}
V.~S. Chen, P.~Varma, R.~Krishna, M.~Bernstein, C.~Re, and L.~Fei-Fei, ``Scene
  graph prediction with limited labels,'' in \emph{ICCV}.\hskip 1em plus 0.5em
  minus 0.4em\relax {IEEE}, 2019, pp. 2580--2590.

\bibitem{Hooper2021cutout}
S.~Hooper, M.~Wornow, H.~S. Ying, H.~Kellman, Peter~andXue, F.~Sala,
  C.~Langlotz, and C.~R{\'e}, ``Cut out the annotator, keep the cutout: better
  segmentation with weak supervision,'' \emph{ICLR}, 2021.

\bibitem{garcia2021distillation}
N.~C. Garcia, S.~A. Bargal, V.~Ablavsky, P.~Morerio, V.~Murino, and
  S.~Sclaroff, ``Distillation multiple choice learning for multimodal action
  recognition,'' in \emph{WACV}, 2021, pp. 2754--2763.

\bibitem{van2020survey}
J.~E. Van~Engelen and H.~H. Hoos, ``A survey on semi-supervised learning,''
  \emph{Machine Learning}, vol. 109, no.~2, pp. 373--440, 2020.

\bibitem{yang2021survey}
X.~Yang, Z.~Song, I.~King, and Z.~Xu, ``A survey on deep semi-supervised
  learning,'' \emph{arXiv preprint arXiv:2103.00550}, 2021.

\bibitem{chen2020train}
M.~F. Chen, D.~Y. Fu, F.~Sala, S.~Wu, R.~T. Mullapudi, F.~Poms, K.~Fatahalian,
  and C.~R\'e, ``Train and you'll miss it: Interactive model iteration with
  weak supervision and pre-trained embeddings,'' \emph{arXiv preprint
  arXiv:2006.15168}, 2020.

\bibitem{fu2020fast}
D.~Fu, M.~Chen, F.~Sala, S.~Hooper, K.~Fatahalian, and C.~R{\'e}, ``Fast and
  three-rious: Speeding up weak supervision with triplet methods,'' in
  \emph{ICML}, vol. 119.\hskip 1em plus 0.5em minus 0.4em\relax PMLR, 2020, pp.
  3280--3291.

\bibitem{guzman2014efficiently}
A.~Guzman-Rivera, P.~Kohli, D.~Batra, and R.~Rutenbar, ``Efficiently enforcing
  diversity in multi-output structured prediction,'' in \emph{Artificial
  Intelligence and Statistics}.\hskip 1em plus 0.5em minus 0.4em\relax PMLR,
  2014, pp. 284--292.

\bibitem{lee2016stochastic}
S.~Lee, S.~Purushwalkam, M.~Cogswell, V.~Ranjan, D.~J. Crandall, and D.~Batra,
  ``Stochastic multiple choice learning for training diverse deep ensembles,''
  in \emph{NeurIPS}, 2016, pp. 2119--2127.

\bibitem{lee2017confident}
K.~Lee, C.~Hwang, K.~Park, and J.~Shin, ``Confident multiple choice learning,''
  in \emph{ICML}, vol.~70.\hskip 1em plus 0.5em minus 0.4em\relax PMLR, 2017,
  pp. 2014--2023.

\bibitem{tian2019versatile}
K.~Tian, Y.~Xu, S.~Zhou, and J.~Guan, ``Versatile multiple choice learning and
  its application to vision computing,'' in \emph{CVPR}.\hskip 1em plus 0.5em
  minus 0.4em\relax IEEE, 2019, pp. 6349--6357.

\bibitem{deng2009imagenet}
J.~Deng, W.~Dong, R.~Socher, L.-J. Li, K.~Li, and L.~Fei-Fei, ``Imagenet: A
  large-scale hierarchical image database,'' in \emph{CVPR}.\hskip 1em plus
  0.5em minus 0.4em\relax IEEE, 2009, pp. 248--255.

\bibitem{lin2014microsoft}
T.-Y. Lin, M.~Maire, S.~Belongie, J.~Hays, P.~Perona, D.~Ramanan,
  P.~Doll{\'a}r, and C.~L. Zitnick, ``Microsoft coco: Common objects in
  context,'' in \emph{ECCV}, vol. 8693.\hskip 1em plus 0.5em minus 0.4em\relax
  Springer, 2014, pp. 740--755.

\bibitem{cordts2016cityscapes}
M.~Cordts, M.~Omran, S.~Ramos, T.~Rehfeld, M.~Enzweiler, R.~Benenson,
  U.~Franke, S.~Roth, and B.~Schiele, ``The cityscapes dataset for semantic
  urban scene understanding,'' in \emph{CVPR}, 2016, pp. 3213--3223.

\bibitem{laine2016temporal}
S.~Laine and T.~Aila, ``Temporal ensembling for semi-supervised learning,'' in
  \emph{ICLR}, 2017.

\bibitem{rasmus2015semi}
A.~Rasmus, H.~Valpola, M.~Honkala, M.~Berglund, and T.~Raiko, ``Semi-supervised
  learning with ladder networks,'' in \emph{NeurIPS}, 2015, pp. 3546--3554.

\bibitem{berthelot2019mixmatch}
D.~Berthelot, N.~Carlini, I.~Goodfellow, N.~Papernot, A.~Oliver, and C.~Raffel,
  ``Mixmatch: A holistic approach to semi-supervised learning,'' in
  \emph{NeurIPS}, 2019, pp. 5050--5060.

\bibitem{tarvainen2017mean}
A.~Tarvainen and H.~Valpola, ``Mean teachers are better role models:
  Weight-averaged consistency targets improve semi-supervised deep learning
  results,'' in \emph{NeurIPS}, 2017, pp. 1195--1204.

\bibitem{miyato2018virtual}
T.~Miyato, S.-i. Maeda, M.~Koyama, and S.~Ishii, ``Virtual adversarial
  training: a regularization method for supervised and semi-supervised
  learning,'' \emph{TPAMI}, vol.~41, no.~8, pp. 1979--1993, 2019.

\bibitem{xie2020unsupervised}
Q.~Xie, Z.~Dai, E.~Hovy, T.~Luong, and Q.~Le, ``Unsupervised data augmentation
  for consistency training,'' in \emph{NeurIPS}, 2020.

\bibitem{berthelot2019remixmatch}
D.~Berthelot, N.~Carlini, E.~D. Cubuk, A.~Kurakin, K.~Sohn, H.~Zhang, and
  C.~Raffel, ``Remixmatch: Semi-supervised learning with distribution alignment
  and augmentation anchoring,'' in \emph{ICLR}, 2020.

\bibitem{han2020unsupervised}
T.~Han, J.~Gao, Y.~Yuan, and Q.~Wang, ``Unsupervised semantic aggregation and
  deformable template matching for semi-supervised learning,'' in
  \emph{NeurIPS}, 2020.

\bibitem{li2020comatch}
J.~Li, C.~Xiong, and S.~C. Hoi, ``Comatch: Semi-supervised learning with
  contrastive graph regularization,'' in \emph{ICCV}, 2021, pp. 9475--9484.

\bibitem{hu2021simple}
Z.~Hu, Z.~Yang, X.~Hu, and R.~Nevatia, ``Simple: Similar pseudo label
  exploitation for semi-supervised classification,'' in \emph{CVPR}, 2021, pp.
  15\,099--15\,108.

\bibitem{zhang2021flexmatch}
B.~Zhang, Y.~Wang, W.~Hou, H.~Wu, J.~Wang, M.~Okumura, and T.~Shinozaki,
  ``Flexmatch: Boosting semi-supervised learning with curriculum pseudo
  labeling,'' \emph{arXiv preprint arXiv:2110.08263}, 2021.

\bibitem{varma2018snuba}
P.~Varma and C.~R{\'e}, ``Snuba: automating weak supervision to label training
  data,'' in \emph{VLDB}, vol.~12, no.~3.\hskip 1em plus 0.5em minus
  0.4em\relax NIH Public Access, 2018, p. 223.

\bibitem{heo2020inspector}
G.~Heo, Y.~Roh, S.~Hwang, D.~Lee, and S.~E. Whang, ``Inspector gadget: A data
  programming-based labeling system for industrial images,'' \emph{Proc. {VLDB}
  Endow.}, vol.~14, no.~1, pp. 28--36, 2020.

\bibitem{pal2018adversarial}
A.~Pal and V.~N. Balasubramanian, ``Adversarial data programming: Using gans to
  relax the bottleneck of curated labeled data,'' in \emph{CVPR}.\hskip 1em
  plus 0.5em minus 0.4em\relax {IEEE}, 2018, pp. 1556--1565.

\bibitem{das2020goggles}
N.~Das, S.~Chaba, R.~Wu, S.~Gandhi, D.~H. Chau, and X.~Chu, ``Goggles:
  Automatic image labeling with affinity coding,'' in \emph{SIGMOD}, 2020, pp.
  1717--1732.

\bibitem{boecking2021interactive}
B.~Boecking, W.~Neiswanger, E.~Xing, and A.~Dubrawski, ``Interactive weak
  supervision: Learning useful heuristics for data labeling,'' \emph{ICLR},
  2021.

\bibitem{varma2017inferring}
P.~Varma, B.~He, P.~Bajaj, I.~Banerjee, N.~Khandwala, D.~L. Rubin, and
  C.~R{\'e}, ``Inferring generative model structure with static analysis,'' in
  \emph{NeurIPS}, vol.~30.\hskip 1em plus 0.5em minus 0.4em\relax NIH Public
  Access, 2017, pp. 240--250.

\bibitem{bach2017learning}
S.~H. Bach, B.~He, A.~Ratner, and C.~R{\'e}, ``Learning the structure of
  generative models without labeled data,'' in \emph{ICML}, vol.~70.\hskip 1em
  plus 0.5em minus 0.4em\relax PMLR, 2017, pp. 273--282.

\bibitem{varma2019learning}
P.~Varma, F.~Sala, A.~He, A.~Ratner, and C.~R{\'e}, ``Learning dependency
  structures for weak supervision models,'' in \emph{ICML}.\hskip 1em plus
  0.5em minus 0.4em\relax PMLR, 2019, pp. 6418--6427.

\bibitem{zhang2021weakly}
L.~Zhang, J.~Ding, Y.~Xu, Y.~Liu, and S.~Zhou, ``Weakly-supervised text
  classification based on keyword graph,'' \emph{arXiv preprint
  arXiv:2110.02591}, 2021.

\bibitem{cubuk2020randaugment}
E.~D. Cubuk, B.~Zoph, J.~Shlens, and Q.~V. Le, ``Randaugment: Practical
  automated data augmentation with a reduced search space,'' in \emph{Workshop
  on CVPR}, 2020, pp. 702--703.

\bibitem{devries2017improved}
T.~DeVries and G.~W. Taylor, ``Improved regularization of convolutional neural
  networks with cutout,'' \emph{arXiv preprint arXiv:1708.04552}, 2017.

\bibitem{zagoruyko2016wide}
S.~Zagoruyko and N.~Komodakis, ``Wide residual networks,'' in
  \emph{BMVC}.\hskip 1em plus 0.5em minus 0.4em\relax BMVC, 2016.

\bibitem{ratner2020snorkel}
A.~Ratner, S.~H. Bach, H.~Ehrenberg, J.~Fries, S.~Wu, and C.~R{\'e}, ``Snorkel:
  Rapid training data creation with weak supervision,'' \emph{{VLDB} J.},
  vol.~29, no.~2, pp. 709--730, 2020.

\bibitem{krizhevsky2009learning}
A.~Krizhevsky, ``Learning multiple layers of features from tiny images,''
  \emph{Master's thesis, University of Tront}, 2009.

\bibitem{netzer2011reading}
Y.~Netzer, T.~Wang, A.~Coates, A.~Bissacco, B.~Wu, and A.~Y. Ng, ``Reading
  digits in natural images with unsupervised feature learning,'' in
  \emph{Workshop on NeurIPS}, 2011.

\bibitem{coates2011analysis}
A.~Coates, A.~Ng, and H.~Lee, ``An analysis of single-layer networks in
  unsupervised feature learning,'' in \emph{Artificial Intelligence and
  Statistics}.\hskip 1em plus 0.5em minus 0.4em\relax JMLR Workshop and
  Conference Proceedings, 2011, pp. 215--223.

\bibitem{jaffe2015estimating}
A.~Jaffe, B.~Nadler, and Y.~Kluger, ``Estimating the accuracies of multiple
  classifiers without labeled data,'' in \emph{Artificial Intelligence and
  Statistics}.\hskip 1em plus 0.5em minus 0.4em\relax PMLR, 2015, pp. 407--415.

\bibitem{platanios2017estimating}
E.~Platanios, H.~Poon, T.~M. Mitchell, and E.~J. Horvitz, ``Estimating accuracy
  from unlabeled data: A probabilistic logic approach,'' \emph{NeurIPS},
  vol.~30, pp. 4361--4370, 2017.

\bibitem{traganitis2018blind}
P.~A. Traganitis, A.~Pages-Zamora, and G.~B. Giannakis, ``Blind multiclass
  ensemble classification,'' \emph{IEEE Transactions on Signal Processing},
  vol.~66, no.~18, pp. 4737--4752, 2018.

\end{thebibliography}
	\bibliographystyle{IEEEtran}
	
}

%%%%%%%%%%%%%%%%%%%%%%%%%%%%%%%%%%%%%%%%%%%%%%%%%%%%%%%%%%%%
%%%%%%%%%%%%%%%%%%%%%%%%%%%%%%%%%%%%%%%%%%%%%%%%%%%%%%%%%%%%
\newpage
\appendix

\section{Appendix}

\subsection{Glossary}
The glossary is given in Table \ref{tab:glossary}.

\begin{table}
\centering
\centering
\caption{Glossary of variables and symbols used in this paper.}
\label{tab:glossary}
\begin{tabular}{ll}
\toprule
Symbol & Used for \\
\midrule
$x_l, y_l$                      & Labeled samples and ground-truth labels      \\
$x_u$                            & Unlabeled samples \\
$x_{\cdot}^w$                          & Weakly augmented (labeled / unlabeled) samples \\
$x_{\cdot}^s$                           & Strongly augmented (labeled / unlabeled) samples \\
$\mathcal{Y}$                & Label space. \emph{e.g.}, in CIFAR-10, $\mathcal{Y}=\{1,2,\cdots, 10\}$ \\
$C$                          & Number of categories. $C=|\mathcal{Y}|$ \\
$f$                              & \tabincell{l}{$\mathbb{R}^{HW\times D}$, feature map before global average pooling of backbone} \\
$f(j)$                           & $\mathbb{R}^{D}$, feature vector at spatial position $j$ of $f$\\
$dis(A, B)$                   & Distance between A and B \\
$c_k$                          & $\mathbb{R}^D$, learnable clustering center of the $k$-th LF \\
${f}_k$                    &$\mathbb{R}^D$, input feature of classifier in the $k$-th LF\\
$\mathcal{F}_k$                  & The classifier head in the $k$-th LF\\
$p_k(y|x)$           & $\mathbb{R}^{C}$, output probability of the $k$-th LF\\
$\tau_k$                        & Specialized category set of the $k$-th LF\\
$\bar{p}_k(y|x)$          & $\mathbb{R}^{|\tau_k| + 1}$, output probability over specialized categories and ``abstention'' option \\
$\hat{y}^k$                    & $=\arg max (\bar{p}_k(y|x))$, predicted label of the $k$-th LF\\
$\hat{\mathbf{y}}$             & $=(\hat{y}^1, \cdots,\hat{y}^K)^\intercal \in\mathbb{R}^K$, vectorized predicted laebls of K LFs\\
$\theta$                      & $\theta \in \mathbb{R}^{K\times |\mathcal{Y}|}$ is the parameters of label model \\
$e_{ky}$                          & $e_{ky}:=exp(\theta_{ky})$,  exponent of parameters $\theta_{ky}$\\
$\phi(y,\hat{y}^k)$             & Potential value with the target label $y$ and  predicted label $\hat{y}^k$\\
$P(y, \hat{\mathbf{y}})$        & Joint distribution between target label $y$ and predicted label $\hat{\mathbf{y}}$ in the label model\\
$Z$                               & Normalizer item of $P(y, \hat{\mathbf{y}})$. $Z=\sum_{y\in\mathcal{Y}}\sum_{\hat{\mathbf{y}}\in \tau}P(y, \hat{\mathbf{y}})$\\
$R(\theta,\hat{\mathbf{y}}_u)$  & Regularizer of unlabeled data in the Label Model\\
$z_i$                             & $z_i=\begin{cases} +1 & y_u=i \\ -1 & y_u\neq i \end{cases}$, latent variable of ground-truth label in the $i$-th one-versus-all task\\
$\hat{z}_i^k$                    & \tabincell{l}{$\hat{z}_i^k=\begin{cases} +1 & \hat{y}^k=i \\ 0 & \hat{y}^k=0 \\ -1 & otherwise \end{cases}$, latent variable of $\hat{y}^k$ in the $i$-th one-versus-all task}\\
$\mathbb{E}[\hat{z}_i^kz_i]$      & Expectation of $\hat{z}_i^kz_i$\\
$\hat{\mathbb{E}}[\hat{z}_i^kz_i]$ & Estimated expectation of $\hat{z}_i^kz_i$ over all unlabeled data without ground-truch $z_i$. \\
$\hat{P}(\cdot)$                   & Probability estimated with the observable data. \\
$\hat{a}_i^k$                     & Precision of the $k$-th LF in the $i$-th one-versus-all classification \\

\bottomrule
\end{tabular}
\end{table}

\subsection{Label Model}
%\textbf{Sensitivity of Initialization.} In this paper,
Label Model in Snorkel~\cite{ratner2020snorkel} is also called as ``Generative Model'', which models and integrates the noisy labels provided by $K$ LFs. In this paper, we suppose that the $K$ LFs are independent. Assuming that $\hat{\mathbf{y}} = (\hat{y}_1, \cdots, \hat{y}_K)^\intercal \in \mathbb{R}^K$ is the vectorized form of the predicted labels from $K$ LFs, and $\hat{y}_k$ is the predicted label of the $k$-th model. For clarity, we denote  $\emptyset$ as the ``abstention'' option in the LFs.
Then, following the definition of Snorkel, the label model in Snorkel can be represented as:
\begin{equation}
    \label{eq:snorkel_label}
    P(y, \hat{\mathbf{y}}) = \frac{1}{Z}\prod_{k=1}^{K}\phi(y, \hat{y}^k),
\end{equation}
with the potential function $\phi(y,\hat{y}^k)$:
\begin{equation}
\label{eq:snorkel_potential_2}
    \phi(y, \hat{y}^k) = \left\{\begin{aligned}
        &exp(\theta_{k1} + \theta_{k2}), &if \; \hat{y}^k = y \\
		&exp(\theta_{k1}), &if \; \hat{y}^k \neq y, \hat{y} \neq \emptyset \\
		&1. &if \; \hat{y}^k=\emptyset\\
	\end{aligned}\right.
\end{equation}
where $\theta \in \mathbb{R}^{2K}$.
It can be observed that the potential function in Eq.~(\ref{eq:snorkel_potential_2}) provides the same values for all target categories $y$. However, each LF in our method specializes in multiple categories with different performance. Thus, we extend the parameters $\theta$ to $K\times C$ to support multi-class classification.  In Eq.~(\ref{eq:snorkel_potential_2}), we set $exp(\theta_{k1} + \theta_{k2})$ to guarantee that the potential with $\hat{y}^k=y$ is larger than that with $\hat{y}^k \neq y$. Similarly, we set $1 + exp(\theta_{ky})$ and $1 / (1 + exp(\theta_{ky}))$. Then, we have the potential function:
\begin{equation}
\label{eq:snorkel_potential_3}
    \phi(y, \hat{y}^k) = \left\{\begin{aligned}
        &1 + exp(\theta_{yk}), &if \; \hat{y} \in \tau_k,\; \hat{y}^k = y \\
		&1 / (1 + exp(\theta_{yk})), &if \; \hat{y} \in \tau_k,\;  \hat{y}^k \neq y \\
		&1. &otherwise\\
	\end{aligned}\right.
\end{equation}

However, the conclusion from \cite{lee2016stochastic, tian2019versatile} tells us that MCL tends to be overconfident for the samples whose ground-truth labels are out of the specialized set. In other words, when one LF is fed with a sample whose ground-truth label is out of the specialized set, it may still produce a label in the specialized set with high confidence. Although the cases of overconfidence decrease a lot due to the introduction of the ``abstention'' option in Step 1, these cases still exist in our framework. Therefore, we introduce the item $exp(\theta_{yk})$ to represent the relationship between the predicted label $\hat{y}^k$ and the target label $y$, even when the target labels conflict with the predicted labels. Based on these considerations, we define the four-part potential function in Eq.~(\ref{eq:potential}.

\subsection{Regularizer}
%\textbf{Regularizer.}
\label{sec:app:reg}
We give the complementary formulation  of $P_\theta(\hat{z}_i^k=z_i|\hat{z}_i^k\neq0)$.
Set $\Phi^k(y, \hat{\mathcal{Y}}) := \sum_{\hat{y}^k \in \hat{\mathcal{Y}}}\phi(y, \hat{y}^k)$ for ease of exposition. We write $P_\theta(\hat{z}_i^k=z_i|\hat{z}_i^k\neq0)$ as follows:
\begin{equation}
	\begin{split}
	P_\theta(\hat{z}_i^k&=z_i|\hat{z}_i^k\neq0) \\ &=\frac{P_\theta(\hat{z}_i^k=z_i=1) + P_\theta(\hat{z}_i^k=z_i=-1)}{P_\theta(\hat{z}_i^k\neq0)} \\
	&=\frac{P_\theta(y=i, \hat{y}^k=y, \hat{y}^k\neq0) + P_\theta(y\neq i, \hat{y}^k\neq i, \hat{y}^k \neq 0)}{P_\theta(\hat{y}^k\neq0)} \\
	&= \frac{\phi(i, i)\prod_{k'\neq k}{\Phi^{k'}(i, \{0\}\cup\tau_{k'})} + \sum_{y\neq i}(\Phi^k(y,\tau_k - \{i\})\prod_{k'\neq k}\Phi^{k'}(y, \{0\}\cup \tau_{k'}))}{\sum_{y\in\mathcal{Y}}\Phi^k(y, \tau_k)\prod_{k'\neq k}\Phi^{k'}(y,  \{0\}\cup \tau_{k'})}. \\
	\end{split}
\end{equation}
As mentioned in our paper, the class-wise accuracy $P_\theta(\hat{z}_i^k=z_i|\hat{z}_i^k=1)$ without negative samples  can be written as:
\begin{equation}
	P_\theta(\hat{z}_i^k=z_i|\hat{z}_i^k=1) = = \frac{\phi(i, i)\prod_{k'\neq k}{\Phi^{k'}(i, \{0\}\cup\tau_{k'})} }{\sum_{y\in\mathcal{Y}}\phi(y, i)\prod_{k'\neq k}\Phi^{k'}(y,  \{0\}\cup \tau_{k'})}.
\end{equation}
However, we can estimate $\hat{P}(\hat{z}_i^k=z_i|\hat{z}_i^k\neq0)$ by Eq.~(\ref{eq:triplet}) directly, but it is more difficult to estimate the class-wise accuracy $\hat{P}(\hat{z}_i^k=z_i|\hat{z}_i^k=1)$.
%\subsection{Barely Supervised Learning}
%The analysis of model on labeled data of various prototypicality are reported in \textbf{Appendix}.

\subsection{Model Analysis}
\begin{figure}
	\includegraphics[width=0.95\textwidth]{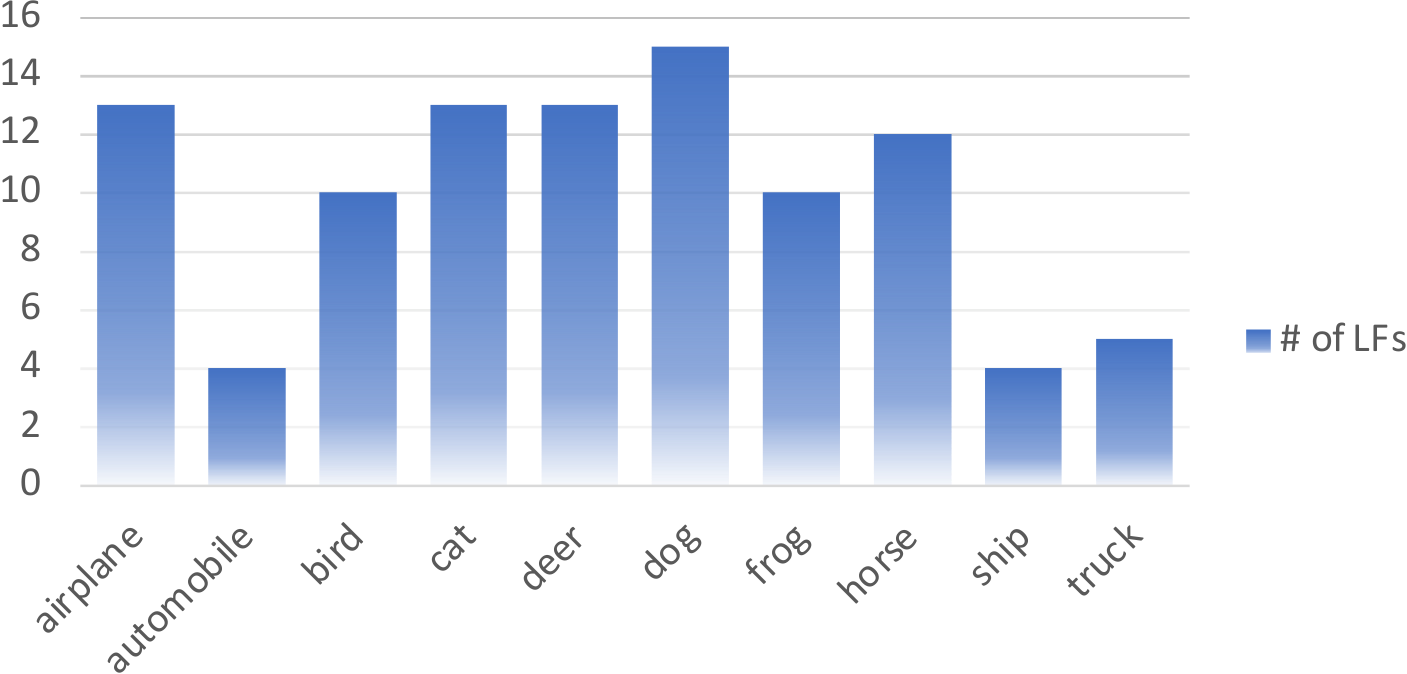}
	\caption{The number of specialized LFs for each category.}
	\label{fig:mv}
\end{figure}
We give an example with 40 labeled samples in Fig.~\ref{fig:mv} to illustrate why Majority Vote falls with $K=50$ and $\rho=0.2$. In the extreme case, category ``automobile'' and category ``ship'' are only specialized by 4 LFs, while 15 LFs specialize in ``dog''. If an image with category ``automobile'' triggers all~(4) true specialized LFs but triggers 40\%~(5) specialized LFs with class ``dog'', Majority Vote would misclassify it into ``dog''. With our label model, we can achieve 95.22\% annotation accuracy in this case.

\subsection{Hyperparameter $\rho$ and $K$}
\label{sec:app:hyper}
We also present the complete experimental results with different $\rho$ and $K$ on CIFAR-10 with 40 labels in Tab.~\ref{tab:full-hyper}.
\begin{table}[]
    \centering
    \caption{Annotation performance for different $\rho$ and $K$ on CIFAR-10 with 40 labels}
    \begin{tabular}{c|cccccc}
         &	0.1 &	0.2	& 0.3&	0.4	& 0.5 &	1.0 \\
         \toprule
         10 &	83.25&	86.42&	88.99&	90.62&	91.28&	92.39\\
        20&	89.16&	91.27&	91.83&	92.31&	92.52&	92.48\\
        30&	91.39&	93.46&	93.54&	93.16&	92.91&	92.41\\
40&	92.53&	93.28&	93.45&	93.24&	92.80&	92.47\\
50&	92.95&	93.82&	93.55&	93.19&	93.07&	92.46\\
60&	93.01&	93.54&	93.11&	92.97&	92.63&	92.43\\
    \end{tabular}
    \label{tab:full-hyper}
\end{table}

%\subsection{Extended Ablations}

%\subsection{Extended Related Work}
%Generally, our method is also related to knowledge distillation.

\end{document}